\documentclass[10pt,twocolumn,letterpaper]{article}

\usepackage[algorithms]{wacv}  
\usepackage{times}
\usepackage{epsfig}
\usepackage{graphicx}
\usepackage{tikz}
\usepackage{booktabs}
\usepackage{dsfont}
\usepackage{multirow}
\usepackage{enumitem}
\usepackage{amsmath}
\usepackage{amssymb}
\usepackage{xcolor}
\usepackage{xparse}
\usepackage{bm}
\usepackage{pifont}
\usepackage{algorithm}
\usepackage{algorithmic}
\usepackage{tcolorbox}
\usepackage{scalerel}
\usepackage{transparent}
\usepackage{textcomp}
\usepackage[acronym]{glossaries}
\graphicspath{ {./figures/} }
\newcommand{\shortsec}[1]{\noindent{\textbf{#1}}}

\newcommand{\best}[1]{\textbf{#1}}
\def\maxvalue(#1,#2){\ifdim#1pt>#2pt #1\else #2\fi}
\newcommand{\plusminus}[1]{\textcolor{white!20!black}{\scaleobj{.8}{ \pm#1}}}
\newcommand{\valdif}[2]{\ensuremath{#1\plusminus{#2}}}
\newcommand{\valmaxmin}[3]{\valdif{#1}{\maxvalue(#2,#3)}}
\newcommand{\valdifb}[2]{\ensuremath{\best{#1}\plusminus{#2}}}
\newcommand{\valmaxminb}[3]{\valdifb{#1}{\maxvalue(#2,#3)}}
\newcommand{\map}{\ensuremath{\text{mAP}}}
\newcommand{\mapfifty}{\ensuremath{\text{mAP}_{50}}}

\newcommand{\iou}{\text{IoU}}
\newcommand{\iouthresh}{\ensuremath{t_{\iou}}}

\newcommand{\detect}[1]{\ensuremath{d_{#1}}}
\newcommand{\detects}{\ensuremath{\mathcal{D}}} 

\newcommand{\categ}[1]{\ensuremath{k_{#1}}}
\newcommand{\bbox}[1]{\ensuremath{b_{#1}}}
\newcommand{\bboxes}[1]{\boldsymbol{\bbox{#1}}}
\newcommand{\conf}[1]{\ensuremath{s_{#1}}} 
\newcommand{\confs}[1]{\boldsymbol{\conf{#1}}}
\newcommand{\confset}{\bm{\mathcal{S}}}
\newcommand{\bboxset}{\bm{\mathcal{B}}}
\newcommand{\probtp}[1]{\ensuremath{\mathds{P}_{#1}}} 
\newcommand{\weights}{\ensuremath{\boldsymbol{\theta}}}
\newcommand{\tpindic}[1]{\ensuremath{\textit{$\tau$}_{#1}}} 
\newcommand{\com}[1]{\hfill\textcolor{white!58!black}{\# #1}}
\newcommand{\ece}{\text{ECE}}
\newcommand{\func}[1]{\ensuremath{\mathit{f}_{#1}}}
\newcommand{\calibfunctrue}{\func{cal}} 
\newcommand{\cmark}{\ding{51}}%
\newcommand{\gcmark}{\textcolor{green}{\cmark}}
\newcommand{\xmark}{\ding{55}}%
\newcommand{\rxmark}{\textcolor{red}{\xmark}}

\newcommand{\nms}{\text{nms}}
\newcommand{\jaccardd}[1]{\ensuremath{j_{#1}}}
\newcommand{\jaccarddsup}[1]{\ensuremath{j^{sup}_{#1}}}
\newcommand{\jaccardv}[1]{\ensuremath{\mathbf{j}_{#1}}}
\newcommand{\jaccardm}{\ensuremath{\mathbf{J}}}
\newcommand{\funcnms}{\func{\nms}}
\newcommand{\nmsthres}{\ensuremath{t_{\nms}}~}
\newcommand{\duplicate}[1]{\ensuremath{\delta_{#1}}}
\newcommand{\nonduplicate}[1]{\Bar{\duplicate{#1}}}

\DeclareMathOperator*{\argmin}{arg\,min}

\newcommand{\bigo}{\mathcal{O}}
\newcommand{\drawbbx}[6]{\draw[#1,thick] (#2,#3) rectangle (#4,#5);
\draw[#1,fill=#1] (#2,#3) rectangle (#2 + 0.094,#3 -0.058);
\node[anchor=north west, text=black, text width=1pt, inner sep=0] at (#2+0.003,#3 -0.003){\footnotesize #6};}
\newcommand{\drawbbxl}[6]{\draw[#1,thick] (#2,#3) rectangle (#4,#5);
\draw[#1,fill=#1] (#2,#3) rectangle (#2 + 0.235,#3 -0.058);
\node[anchor=north west, text=black, text width=1pt, inner sep=0] at (#2+0.007,#3 -0.003){\footnotesize #6};}
\newcommand{\bxarrow}[3]{
\draw [-stealth,very thick](#3) [left] -- (#1 + 0.095,#2 -0.035);}

\newcommand{\txtarrow}[6]{\node (#6) at (#3,#4) [rectangle,draw,fill=white,minimum size=1.4em,label={center:\footnotesize#5}] {\footnotesize#5};
\bxarrow{#1}{#2}{#6}}

\newacronym{ece}{ECE}{Expected Calibration Error}
\newacronym{ace}{ACE}{Adaptive Calibration Error}
\newacronym{sce}{SCE}{Static Calibration Error}
\newacronym{nll}{NLL}{negative log likelihood}
\newacronym{mle}{MLE}{maximum likelihood estimation}
\newacronym{coco}{COCO}{Common Objects in Context dataset~\cite{lin2015mscoco}}
\newacronym{tta}{TTA}{Test Time Augmentation}
\newacronym{nms}{NMS}{Non-Maximum Suppression}
\newacronym{tp}{TP}{True Positive}
\newacronym{iou}{IoU}{Intersection over Union}
\newacronym{wbf}{wbf}{Weighted Box Fusion \cite{solovyev2021weightedboxfusion}}

\usepackage[pagebackref=true,breaklinks=true,letterpaper=true,colorlinks,bookmarks=false]{hyperref}
\usepackage[capitalize]{cleveref}
\crefname{section}{Sec.}{Secs.}
\Crefname{section}{Section}{Sections}
\Crefname{table}{Table}{Tables}
\crefname{table}{Tab.}{Tabs.}

\begin{document}

\newcommand{\method}{\iou-aware calibration}
\title{Do We Still Need Non-Maximum Suppression? Accurate Confidence Estimates and Implicit Duplication Modeling with IoU-Aware Calibration}
\author{Johannes Gilg \qquad Torben Teepe \qquad Fabian Herzog  \qquad Philipp Wolters \qquad Gerhard Rigoll \vspace{5pt}\\
Technical University Munich \vspace{5pt} \\
{\tt\small Johannes.Gilg@TUM.de}
}

\maketitle

\begin{abstract}
Object detectors are at the heart of many semi- and fully autonomous decision systems and are poised to become even more indispensable. They are, however, still lacking in accessibility and can sometimes produce unreliable predictions. Especially concerning in this regard are the---essentially hand-crafted---non-maximum suppression algorithms that lead to an obfuscated prediction process and biased confidence estimates. We show that we can eliminate classic NMS-style post-processing by using IoU-aware calibration. IoU-aware calibration is a conditional Beta calibration; this makes it parallelizable with no hyper-parameters. Instead of arbitrary cutoffs or discounts, it implicitly accounts for the likelihood of each detection being a duplicate and adjusts the confidence score accordingly, resulting in empirically based precision estimates for each detection. Our extensive experiments on diverse detection architectures show that the proposed IoU-aware calibration can successfully model duplicate detections and improve calibration. Compared to the standard sequential NMS and calibration approach, our joint modeling can deliver performance gains over the best NMS-based alternative while producing consistently better-calibrated confidence predictions with less complexity. The \hyperlink{https://github.com/Blueblue4/IoU-AwareCalibration}{code} for all our experiments is publicly available. 

\end{abstract}

\section{Introduction}
\begin{figure}
\begin{center}
 \resizebox{1.05\columnwidth}{!}{%
        \input{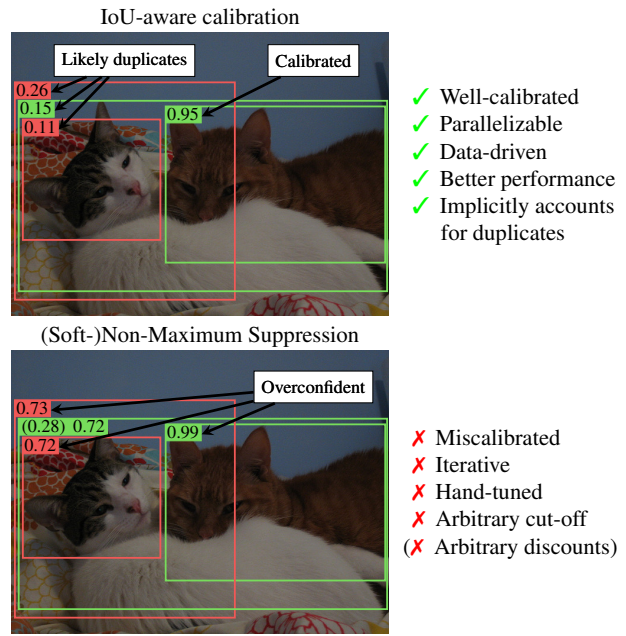}%
         }
\end{center}
   \caption{\textbf{Visualization of detections after \method~and (Soft-)NMS:} with comparison of their respective strengths (\gcmark) and weaknesses (\rxmark).}
   \vspace{-1em}
\label{fig:visual_example}
\end{figure}
Object detectors are indispensable in many fields today, including computer vision, robotics, and autonomous systems. They locate and identify objects in images or videos, allowing for object recognition, tracking, and scene understanding. Object detectors are also at the heart of recent advances in driver-assistance systems, surveillance systems, and augmented reality. With the increasing demand for advanced technologies in areas such as security, industrial automation, and even retail, object detectors are poised to play an even more important role in the future. \\
The proliferation of object detectors appears inevitable, but some hurdles remain. Users will likely include more non-experts, so transparency and accessibility---where today's detectors are still lacking---will be essential. Detectors will also need to be reliable and explainable, as deployments at-scale can amplify existing issues and biases. Researchers and practitioners need to examine current object detection pipelines closely and address their shortcomings in these critical areas of reliability, explainability, and biases. \\ 
A particularly troublesome part of almost all object detector's post-processing pipelines is \gls{nms}. It is a hand-crafted algorithm that is intended to reduce duplicate detections of a single object. For \gls{nms} to work correctly, designers need to define a specific threshold that controls which detections are considered redundant when they overlap. Soft-\gls{nms}---the most popular improvement on the traditional \gls{nms} algorithm---introduces a different hyper-parameter that determines the degree to which overlapping detections' confidence scores are discounted via a Gaussian~\cite{bodla2017soft_nms}. This rather unintuitive hyper-parameter can have a high performance and reliability impact and needs to be chosen carefully.  \\ 
Using an algorithm designed for one use case in another domain or even just adapting it to frequent domain shifts can be cumbersome. It requires knowledge about the target domain, an understanding of the workings of the used \gls{nms} algorithms parameters, and how they influence the detectors' predictions. Alternatively, the parameter space can be searched for performance-maximizing settings, but this still leaves the user lacking an even superficial understanding of their implications.  \\ 
Precise probability assessments are vital for automated decision-making, ensuring accurate and dependable outcomes and proper risk evaluation. 
The discontinuities introduced by \gls{nms} and the confidence discounting by the soft-\gls{nms} distort the distribution of detection confidences that is supposed to reflect accurate probabilities. \Gls{nms} has been shown to impact the calibration of confidence predictions \cite{schwaiger2021nms_miscal}, that are often already miscalibrated to begin with~\cite{guo17cal_temscale}. When the confidence values do not reflect the empiric probability of object presence, they devolve into an inscrutable relative-likelihood ranking between different detections that are largely useless for automated systems.  \\
We propose to map the issue of removing duplicate detections into the confidence score as one unified decision criteria with our data-driven \method. We show that our proposed \method~can implicitly account for the likelihood of each detection being a duplicate of another detection and produce well-calibrated confidence estimates, eliminating the need for other post-processing steps like \gls{nms} and ``normal" confidence calibration. \vspace{0.2em}\\
\shortsec{\method~offers several advantages:}
\vspace{-0.1em}
\begin{enumerate}[nosep]
    \item it is a black-box calibration method, \ie can be applied to any object detector as a post-processing,
    \item it jointly tackles the problem of \gls{nms} and confidence calibration by implicitly modeling the probability of duplicate detections, making \gls{nms} obsolete and producing more reliable confidence predictions,
    \item it can be vectorized as it does not require the iterative $\bigo(N^2)$ computations of soft- and greedy-\gls{nms},
    \item it has no hyper-parameters that need to be manually tuned,
    \item it produces detectors with similar or better performance compared to the best \gls{nms} with better calibration on a wide range of tested detectors.

\end{enumerate}

\section{Related Work}
\shortsec{Non-maximum suppression.}
The numerous variations of the standard \gls{nms} algorithm can be broadly categorized into \gls{nms}-improvements, task-specific adaptations, and proposed modifications to detection architectures. Improvements are \eg, the soft-\gls{nms} that utilizes discounts instead of hard cut-offs \cite{bodla2017soft_nms} and \gls{wbf} that aims to fuse overlapping detections but also algorithmic improvements in processing speed via matrix-operations \cite{bolya2019fast_nms, wang2020solov2_matrix_nms} or spacial-priors to avoid unnecessary computations \cite{tripathi2020asapnms}. \Gls{nms}-adaptations to specific domains include modifications specifically for vehicle-detections \cite{cui2019gaussian_worse} and a differentiable \gls{nms} for 3D-object detection \cite{Kumar2021GrooMeD_NMS}. 
Numerous approaches require specific object detection architectures and modifications thereof \cite{jiang2018acquisition,Tychsen-Smith2018fitness_nms,he2018softer,Liu2019adaptive_nms,zhao2022d_nms, zhou2020noh_nms}. We do not detail these approaches as we aim for a black-box post-processing replacement for \gls{nms} that can be applied to object detectors regardless of architecture. 
\\
A different line of research focuses on removing the need for \gls{nms} altogether by utilizing one-to-one assignment of detection proposals to ground truth objects \cite{carion2020endtoend_detr,sun2021sparsercnn}. This approach can have some drawbacks, as it requires many times more training iterations and has lower performance than comparable architectures that rely on \gls{nms} \cite{chen2022groupdetr,ouyang2022nms}. This line of research is orthogonal to ours, as we show with our experiments on \gls{nms}-less architectures (see \cref{sec:experiments}).
\vspace{0.5em}\\
\shortsec{Confidence calibration.}
Since the findings of Guo \etal---deep-learning models are often highly miscalibrated---there has been a renewed interest in improving on the classic confidence calibration methods like Histogram binning \cite{zadrozny2001histbin} and
Platt Scaling \cite{platt1999scalingprobabilistic,guo17cal_temscale}.
Most closely related to our work are the improvements on Platt Scaling, like the more expressive Beta \cite{kull2017betacal}, the respective multi-class adaptations temperature scaling \cite{guo17cal_temscale} and Dirichlet \cite{kull2019dirichlet}. We claim no originality of the calibration function and rely on the previously introduced multivariate calibration methods for object detectors \cite{kuppers2020multivariate}. Our approach can be improved with further advances in multivariate calibration methods. \\
%
\section{Background}
Object detectors are almost always trained with a many-to-one assignment of detection proposals to ground truth objects. This approach increases the detection performance \cite{chen2022groupdetr,ouyang2022nms,ge2021yolox}, as it can alleviate the foreground-background class imbalance problem of detectors---there are countless non-objects in every image and only a limited number of detectable objects \cite{lin2017focal}. This training strategy, by definition, incentivizes multiple detections per object. These excess detections then have to be filtered out during inference. 
\subsection{Non-Maximum Suppression} 
The combinatorial optimization problem of deciding which detections to keep and which to drop is usually still solved by using either \gls{nms} or soft-\gls{nms}~\cite{bodla2017soft_nms} (see \cref{alg:nms}). Soft-\gls{nms} and the standard \gls{nms} algorithm rely on the premise that duplicate detections are assumed to have similar scales and are highly overlapping. This ``closeness" of two detections \detect{a} and \detect{b} is quantified via the Jaccard index, also termed \gls{iou}, of their respective bounding-boxes \bbox{a} and \bbox{b}:
\begin{equation}
	\iou(\bbox{a}, \bbox{b}) = \frac{\bbox{a} \cap \bbox{b}}{\bbox{a} \cup \bbox{b}}.
    \label{eq:iou}
\end{equation}
It should not be confused with the Jaccard-distance which is $1 - \iou(\bbox{a}, \bbox{b})$.
The second shared assumption is that a higher predicted confidence \conf{i} \textgreater \conf{j} indicates a higher likelihood for the detection \detect{i} to be a \gls{tp} detection than detection \detect{j}. Thus, both algorithms greedily evaluate all $N$ detections \detects{} from highest to lowest confidence \conf{}. At detection $i$ the confidence \conf{i} is updated according to:
\begin{equation}
	\conf{i} = \conf{i} \cdot \funcnms(\bbox{i}, \bbox{j}), \qquad \forall j \hspace{0.5em} | \hspace{0.5em} \conf{j} < \conf{i}.
\label{eq:discount_conf}
\end{equation}
\Gls{nms} uses an arbitrarily defined sharp cutoff threshold \nmsthres to classify detections as duplicates of a more confident detection and discounts them completely:
\begin{equation}
  \funcnms(\bbox{a}, \bbox{b}, \nmsthres)=\left\{
        \begin{array}{ll}
            1,
            &\iou(\bbox{a}, \bbox{b}) < \nmsthres 
            \\
            0,
            &\iou(\bbox{a}, \bbox{b}) \geq \nmsthres
        \end{array}
        ,
        \right.
    \label{eq:greedy-nms}
\end{equation}
while soft-\gls{nms} only discounts the confidence of overlapping detections according to their overlap, usually by a Gaussian:
\begin{equation}
        \funcnms(\bbox{a}, \bbox{b}, \sigma) = e^{-\frac{\iou(\bbox{a}, \bbox{b})^2}{\sigma}}, 
    \label{eq:soft-nms}
\end{equation}
with a separate hyper-parameter $\sigma$. It was designed with the assumption that detections with an overlapping detection that is more confident can still be \gls{tp} detections; they are just less likely to be so.
\begin{algorithm}
    \caption{Greedy \gls{nms} pseudo code} \label{alg:nms}  
    
    \begin{algorithmic}[1]
    \REQUIRE{
        $ $\\
        $
            \bboxset = \{ \bbox{1},\dots,\bbox{N}\} 
        $ \com{Bounding-boxes}
        \newline
        $
            \confset = \{\conf{1},\dots,\conf{N}\}
        $ \com{ Confidences} \\
        $h$ \com{Hyperparmeter: \ensuremath{\sigma} or  \ensuremath{\nmsthres}}
        }\vspace{0.5em}
    
        \STATE $\bm{\mathcal{D}}  = \{\}; \bm{\mathcal{C}}  = \{\}$ \com{output Boxes and Confidences} \\
        \WHILE{$\confset \neq \varnothing$}{
            \STATE $i = \text{argmax} (\confset)$\\
            \STATE{$\confset = \confset \setminus \conf{i};
             \bboxset = \bboxset \setminus \bbox{i}$\\}
             \STATE{$\bm{\mathcal{C}} = \bm{\mathcal{C}} \cup \conf{i};
             \bm{\mathcal{D}} = \bm{\mathcal{D}} \cup \bbox{i}$\\}
            \FOR{$\bbox{j} \in \bboxset$}{
                \STATE{
                        $\conf{j} = \conf{j}\cdot \funcnms(\conf{j}, \conf{i}, h)$ \com{\cref{eq:greedy-nms} or \cref{eq:soft-nms}}
                }
                \IF{\conf{j} == 0}{
                \STATE{$\confset = \confset \setminus \conf{j};
             \bboxset = \bboxset \setminus \bbox{j}$\com{optional} \\ }
             }
             \ENDIF
            }
            \ENDFOR
        }
        \ENDWHILE
        \RETURN{$\bm{\mathcal{D}}, \bm{\mathcal{C}}$}
    \end{algorithmic}
    
\end{algorithm}
\subsection{Object Detector Calibration}
Suppose a model predicts the presence of an object with its detection \detect{} with confidence \conf{}, bounding box \bbox{}, and category \categ{}. The model is \emph{calibrated} if \conf{i} is always equal to the probability of it being a \gls{tp} detection (\tpindic{}=1), \ie, its precision at \conf{i}:
\begin{equation}
    \probtp{}(\tpindic{} = 1 | \conf{} = \conf{i} ) = \conf{i} ,
    \label{eq:calibrated}
\end{equation}
for any $\conf{i} \in [0,1]$. The concept of a \gls{tp} detection in the object detection setting is more involved than in the classification setting, where the only requirement is that the predicted category \categ{} matches the ground truth label. For a detection to be evaluated as a \gls{tp} ($\tpindic{}=1$) it needs to have the highest confidence \conf{} among the set of all detections \detects{} that have the correct category \categ{} and a bounding-box \bbox{} that sufficiently overlaps a ground truth object. The sufficient overlap is determined according to a threshold \iouthresh~on the \iou~ (\cref{eq:iou}) of the ground truth and detected object. This introduces some ambiguity into the concept of a \gls{tp} detection as to what a ``good" value for \iouthresh{} is. In the performance evaluation metric \map{} this is resolved by evaluating over a range of thresholds $\iouthresh \in [0.5, 0.55, ..., 0.95]$ \cite{lin2015mscoco}. A binary label is usually required for the confidence calibration, so we use $\iouthresh=0.5$, same as \mapfifty{}, unless specified otherwise \cite{kuppers2020multivariate}.
\vspace{-2em}
\subsubsection{Measuring Calibration Error}
The \gls{ece} tries to capture the expected difference of the left- and right-hand sides of \cref{eq:calibrated} over some data distribution and the whole confidence interval. To make this evaluation practical certain simplifications are used. The confidences \conf{} can take any value on the interval [0,1] and have to be discretized by splitting the interval into a fixed number ($B$), usually $B$=10, of bins $b$, each containing $n_b$ detections. The precision for each bin $\probtp{}(\tpindic{} = 1 | \conf{} \in b)$ is evaluated over $N$ detections on a labeled hold-out set. The result can be visualized in a reliability diagram \cite{degroot1983comparison_reliabilitydiag} or averaged to produce the detection \gls{ece}-metric~\cite{guo17cal_temscale,kuppers2020multivariate}:
\begin{equation}
    \ece = \sum^{B}_{b=1} \frac{n_b}{N} |\mathit{prec}(b) - \mathit{conf}(b)| .
    \label{eq:ece}
\end{equation}
The standard \gls{ece} was shown to have numerous pathologies. As a remedy, Nixon \etal introduced two variations on the \gls{ece}. The \gls{sce} aims to resolve the problem of class dependency by calculating the \gls{ece} for each class separately and then averaging the results. The second metric, \gls{ace}, uses an adaptive binning scheme. The confidences of object detectors are usually not uniformly distributed but are clustered more densely at the extreme values close to 0 and 1. To account for this imbalance, \gls{ace} uses spaced bins, so each bin contains approximately the same number of detections ~\cite{nixon2019measuring_cal}. In this work, all of the calibration-metrics, \ie  \gls{ece}, \gls{ace}, and \gls{sce}, are stated in percent unless otherwise indicated.
\\
Proper scoring rules like \gls{nll} are often used to fit and validate calibration methods \cite{kull2017betacal}. \gls{nll} is minimized only if there is no miscalibration but does not directly capture the calibration error \cite{hastie2009elements,gneiting2007strictly}.
\subsubsection{Confidence Calibration}
Modern neural networks are usually not well calibrated \cite{guo17cal_temscale}, and neither are modern deep-learning based object detectors \cite{neumann2018relaxed}. The goal of model-agnostic (black-box) confidence calibration is to adjust the confidence estimates produced by a statistical model to better reflect the actual probability of the predicted outcomes using some mapping $\conf{i} \mapsto \calibfunctrue(\conf{i})$, that is inspired by probability distributions. In the context of object detection:
\begin{equation}
    \probtp{}(\tpindic{} = 1 | \conf{} = \conf{i} ) \overset{!}{=} \calibfunctrue(\conf{i}, \weights).
    \label{eq:calibrating}
\end{equation}
According to this notation, the calibration functions are parametrized with $\weights$. There are also non-parametric calibration functions like Histogram Binning \cite{zadrozny2001histbin}, Isotonic Regression \cite{chakravarti1989isotonic}, and Bayesian Binning \cite{naeini2015bayesbin}, which we omit here because they pose some unique problems in the multivariate calibration setting~\cite{gilg2023box}. Common parametrized calibration methods include Platt scaling \cite{platt1999scalingprobabilistic}, Beta calibration \cite{kull2017betacal}, and Temperature scaling \cite{guo17cal_temscale}. Note that most parametrized calibration functions are defined for logit outputs not the final confidence values \conf{i} we use for simplicity, but a conversion is possible. 
\\
The parameters of the calibration function can not be directly optimized on their defined objective shown in \cref{eq:calibrating}. They are instead optimized according to some scoring rule, like the \gls{nll}-loss or the Brier Score \cite{brier1950verification} resulting in an optimization criteria like:
\begin{equation}
     \argmin_{\weights} \sum_{i} (\tpindic{i} - \calibfunctrue(\conf{i}, \weights))^2 .
    \label{eq:calibrate_obj}
\end{equation}
In this form, the calibration can be posed as an optimization of a logistic regression problem and can be solved using gradient-based optimization procedures. 
\subsubsection{Multivariate Confidence Calibration} 
Confidence calibration can be further extended to model bi- or multivariate probability distributions to apply them to multi-output models. Object detectors produce additional regression outputs, \ie, their bounding-box predictions, that can be used for multivariate calibration. The objective of a multivariate calibration of an object detector \eg:
\begin{equation}
    \probtp{}(\tpindic{} = 1 | \conf{} = \conf{i},  \bbox{} = \bbox{i}) \overset{!}{=} \calibfunctrue(\conf{i}, \bbox{i}, \weights),
    \label{eq:multivar_calibrating}
\end{equation}
the motivation is to remove any conditional confidence biases with regard to object positions or size. Multivariate calibration functions can be derived from their univariate counterparts \cite{olkin2015constructions,libby1982multivariate}. If we model the conditioned-on variables---in this case \conf{} and \bbox{}---as independent of each other, the modification of the calibration function is a relatively straightforward vectorized version of its univariate counterpart. When we reasonably assume that the variates are dependent on each other, we also need to model the covariance matrices. We use the implementation by Küppers \etal, which introduced conditionally dependent and independent multivariate adaptations for the logistic and Beta calibration in their multivariate calibration framework for object detectors \cite{kuppers2020multivariate}. 
\\
Unlike the uni-variate calibration, the multivariate confidence calibration can impact a model's performance if there is any conditional bias regarding the additional model outputs \cite{gilg2023box}.
%
\section{\method} 
Object detectors are an integral component of the vision stack in automated decision-making systems. To ensure trust in these systems, it is crucial that object detectors reliably produce predictions with precise probability estimates. The reliability of these predictions is essential in making informed decisions and achieving dependable outcomes in automated systems. In stark contrast to this goal, the hand-crafted discounting and suppression of the ubiquitous \gls{nms}-methods obscure the meaning of detector outputs.
\\
The prevalent training paradigm of many-to-one detection-to-object assignment---and, to a lesser extent, the one-to-one assignment---makes the detectors produce overlapping duplicate detections for individual objects. These confident duplicate detections can severely impact the detector's performance and are, therefore, usually discarded or heavily discounted, depending on their overlap with other detections by heuristic-based algorithms.
\\
We argue that if the confidence of predictions properly accounted for the likelihood of the detections being duplicates, we could eliminate the \gls{nms} post-processing altogether. As likely-duplicated detections would have very low confidence---reflecting their low overall probability of being the true detection.
Therefore, we aim to develop a data-driven approach that produces transparent decision criteria to improve the accuracy and reliability of object detection. This approach must ensure that the empirical probability of overlapping detections being duplicates is properly estimated, leading to well-calibrated reliable confidence predictions. 
\\
If we allow for the implicit assumption of most \gls{nms} approaches, then the likelihood of a detection being a duplicate \duplicate{i} is dependent on its confidence \conf{i}, and the \gls{iou} with all other detections of the same class, which we represent via the Jaccard-distance vector \jaccardv{i} which is $[1-\gls{iou}(\bbox{i},\bbox{1}), 1-\gls{iou}(\bbox{i},\bbox{2}), ... 1-\gls{iou}(\bbox{i},\bbox{N})]^T$. If we want to include the  explicitly calculated probabilities of a detection being a duplicate in the calibration definition of \cref{eq:calibrated}, it results in: 
\begin{equation}
    \conf{i} = \underbrace{\probtp{}(\tpindic{} = 1 | \duplicate{i}) \probtp{}(\duplicate{i}| \conf{i},  \jaccardv{i})}_{=0} +  \underbrace{\probtp{}(\tpindic{} = 1 | \nonduplicate{i}) \probtp{}(\nonduplicate{i}| \conf{i},  \jaccardv{i})}_{\probtp{}(\tpindic{} = 1 | \conf{i},  \jaccardv{i})}.
    \label{eq:objective_duplicate_probs}
\end{equation}
This requires us to explicitly determine the conditional probability of a detection not being a duplicate $(\probtp{}(\nonduplicate{i}| \conf{i},  \jaccardv{i}))$ and the conditional probability of a  non-duplicate detection being a correct, i.e. \gls{tp}, detection $(\probtp{}(\tpindic{} = 1 | \nonduplicate{i}))$. We can simplify this by empirically determining $\probtp{}(\tpindic{} = 1 | \conf{i},  \jaccardv{i})$, thereby skipping the explicit calculation of the non-duplicate probability. This approach transforms the problem into a bi-variate confidence calibration (as defined in \cref{eq:multivar_calibrating}). 
\\
The conditioning on \jaccardv{i} poses a practical problem as it is not a single parameter but a vector with a length that varies with the number of detections. The \gls{nms} methods solve this problem by iterating over all the box-parings of boxes with higher confidence (see \cref{eq:discount_conf}), but we can also use a basic summary statistic, like the minimum, of the Jaccard-distances similar to the approach proposed for the Matrix Non-Maximum Suppression~\cite{wang2020solov2_matrix_nms}. These summary statistics can easily be calculated in parallel over the whole Jaccard-Distance-Matrix of (1 - \gls{iou}) values between all detections:
\begin{equation}
    \jaccardd{i, min} = \min_{\forall \conf{k} > \conf{i}} \{1 - \iou(\bbox{i},\bbox{k})\},
    \label{eq:jaccard_distancen_min}
\end{equation}
removing the need for the serial computation with $\bigo(N^2)$ complexity of the \gls{nms} algorithm for $N$ detections (compare \cref{alg:nms} to \cref{alg:algorithm}). We will use $\jaccardd{i, min}$ as a proxy for \jaccardv{i} unless specified otherwise and ablate this choice later (see \cref{subsec:ablations}). From these design choices, our optimization objective becomes:
\begin{equation}
     \argmin_{\weights} \sum_{i} (\tpindic{i} - \calibfunctrue(\conf{i}, \jaccardd{i, min}, \weights))^2 .
    \label{eq:cond_calibrate_obj}
\end{equation}
For the calibration function, we choose the Bi-variate conditional Beta calibration, as it is more expressive than the Logistic calibration; it produces the identity function if the predictions of a model are already well calibrated, at the cost of a few more parameters \cite{kull2017betacal}. We assume conditional dependence between our overlap-proxy (\jaccardd{i}) and the confidence \conf{i} because we calculate the minimum Jaccard-Distance only from the more confident detections, \ie, the detections that would ``suppress" detection \detect{i} in the classical \gls{nms} setting. This already introduces an indirect dependence between the two variables, which we want to account for in our modeling, but we will also ablate this choice. \\
With this approach, we can now calibrate detections conditioned on a summary statistic of the Jaccard-Distances to the other detections, thereby implicitly accounting for the likelihood of each detection being a duplicate. This should help object detectors reliably produce predictions with precise probability estimates and remove the need for \gls{nms} post-processing.
\begin{algorithm}[t!]
    \caption{Vectorized implementation of \method~in pseudo code} \label{alg:algorithm}  

    \begin{algorithmic}[1]
    \REQUIRE{$ $
        \\
        $
            \bboxes{} = [\bbox{1},\dots,\bbox{N}] 
        $ \com{ Bounding-boxes, shape: [N, 4]}
        \newline
        $
            \confs{} = [\conf{1},\dots,\conf{N}]
        $ \com{ Confidences, shape: [N]} \\
        }\vspace{0.5em}
    
        \STATE $\bboxes{}, \confs{}  = \text{sort\_descending}((\bboxes{}, \confs{}), \text{by=} \confs{})$ \com{[N, 4], [N]}  \\
        \STATE $\jaccardm = \text{ones}(N,N) - \iou(\bboxes{}, \bboxes{}^T) $ \com{Jaccard-dists.,  [N, N]}  \\
        \STATE $\jaccardm = \jaccardm \cdot \text{lower\_triangular(N,-1)}$ \com{Mask sup.,  [N, N]}\\ 
        \STATE$ \jaccardv{\min} = \text{min}(\jaccardm , \text{axis=1})$ \com{\cf \cref{eq:jaccard_distancen_min}, [N]} \\
        \STATE $\confs{} = \text{Conditional\_Beta\_Calibration}(\confs{}, \jaccardv{\min}) $ \com{[N]}\\
    \RETURN{$\bboxes{}, \confs{} $ \com{ [N, 4], [N]}}
    \end{algorithmic}
    
\end{algorithm}
\section{Experiments}\label{sec:experiments}
\begin{figure}[t!]
    \centering
    \resizebox{0.99\columnwidth}{!}{%
    \includegraphics{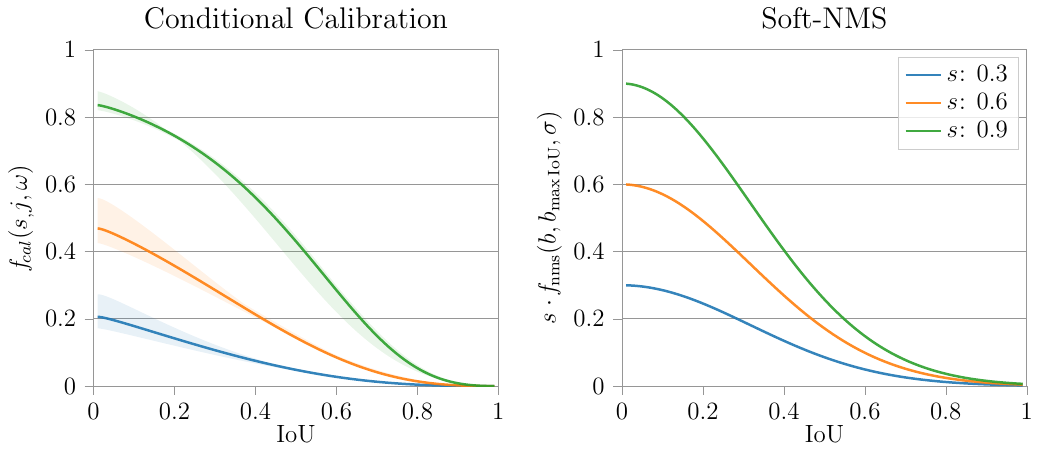}
    }
    \caption{\textbf{Comparison of proposed \method~and soft-\gls{nms}.} Shows how confidence of detections is adjusted, depending on the \iou{} with a more confident detection with three different initial confidences \conf{}. Confidence intervals in lighter colours. Note how the conditional calibration can account for the different confidences with slightly varying curve shape and discounting factor. We show the best-performing soft-\gls{nms} with $\sigma$=0.2.} 
    \label{fig:compare_cc_softnms}
\end{figure}

%
We now verify the theoretical justifications for our \method~by conducting detailed experiments and analyze individual design choices through ablations. 
\\
\textbf{The setup.}  For our initial experiments we use a two-stage architecture, as the region-proposal mechanism makes them produce more duplicate detections, making the effects more measurable. We use a modern pre-trained version of the highly-popular Faster-RCNN \cite{Ren2017fasterrcnn} with a Resnet50 \cite{he2016resnet} backbone and a FPN-Neck \cite{lin2017featurefpn}, trained with multi-scale data augmentation for 36 epochs on \gls{coco}. We remove the \gls{nms} post-processing and limit the maximum number of detections per image to 400 to allow for the additional duplicate detections. The number of detections is later limited to the standard 100 per image for evaluation. We use the multivariate calibration framework for object detectors \cite{kuppers2020multivariate} for the implementation of the Logistic and Beta calibration and their optimization. As usual for calibration methods, we split the \gls{coco} validation set, \texttt{val2017}, into a fitting and an evaluation sub-set. Specifically, we evaluate across 10 random 60:40 train:test image-wise splits and report the mean performance with the maximum difference from it. To evaluate if a detection is a \gls{tp} (\tpindic{} = 1) we use the official \gls{coco}-evaluation script and a \iou-threshold of \iouthresh=0.5 \cite{lin2015mscoco}. \\
For \gls{nms}, the reported metrics are evaluated on the \emph{same splits}, and the hyper-parameters for the different methods are determined by a parameter grid-search on the whole \texttt{val2017}, reflecting the \emph{best possible performance} achievable by \gls{nms}. When only one \gls{nms} method is shown, we report the best performing out of Gaussian soft-\gls{nms}, standard \gls{nms}, and \gls{wbf}.
\\
We apply our proposed \method~in the described cross-validation scheme. We want to test if it can implicitly account for the likelihood of each detection being a duplicate. If our method can capture it, we should see a similar performance increase to a \gls{nms}-based post-processing. We verify this via the performance-metrics \map{} and \mapfifty. Additionally, we aim for the detector to reliably produce predictions with precise probability estimates, which we capture with the calibration-metrics: \gls{ece},
\gls{ace}, \gls{sce} and to a lesser extent with the \gls{nll}. Our method tightly couples the two issues of performance and calibration. Since a performance increase means our method is able to implicitly capture at least the relative likelihood of duplication, the explicitly optimized-for empiric probability assessment $(\probtp{}(\tpindic{} = 1 | \conf{i}, \jaccardv{i}))$ is likely also a well-calibrated prediction (see \cref{eq:objective_duplicate_probs}). \\
Since \gls{nms} has been shown to cause severe miscalibration \cite{schwaiger2021nms_miscal}, we add a uni-variate confidence calibration with the Beta calibration for a better calibration-metrics baseline.
\\
\shortsec{Initial results.} The results are shown in \cref{tab:baseline}. As expected, our \method~produces well-calibrated detections. The significant improvement over the best calibrated \gls{nms} results provides evidence that the implicit modeling of the probability of duplicate detections is indeed a critical necessity to produce precise probability estimates. 
The \method~also performs very well on the performance-metrics. It even outperforms the best \gls{nms}, in this case soft-\gls{nms} by a very slight margin and \gls{nms} by 0.6 \map. This is especially surprising since our method relies only on the Jaccard distance to the most overlapping other detection. In contrast, the iterative \gls{nms}-methods consider all overlapping detections with their iterative approach. 
\\
\shortsec{What does \method~do?} Intrigued by the excellent results, we examine what the learned $\calibfunctrue$ does (see \cref{fig:compare_cc_softnms} left) and compare it to the Gaussian soft-\gls{nms} (see \cref{fig:compare_cc_softnms} right). Visual inspection reveals that the \method~adjusts the shape of the confidence mapping not only by the amount of overlap (in \iou{}) but also by the initial confidence value. While the curve approximates a Gaussian for \conf{}=0.9, it resembles to a linear decay for \conf{}=0.3. The confidence discounting for high \iou{}s is also much reduced, especially for initially high confidence values, compared to the soft-\gls{nms}. The soft-\gls{nms} is applied iteratively (see \cref{eq:discount_conf}) for each more-confident detection, so detections with multiple overlapping higher-confident detections would have their confidence discounted even stronger.  
\begin{table}[t!] 
        \centering{%
        \resizebox{\columnwidth}{!}{%
        \begin{tabular}{llcccccc} 
\toprule
\multicolumn{2}{l}{Post-Processing} & \multicolumn{2}{c}{Performance-Metrics} & \multicolumn{4}{c}{Calibration-Metrics}  \\
                \gls{nms} & Calibration &          \map \textuparrow &                          \mapfifty \textuparrow &                           ECE\textdownarrow &                           
                ACE\textdownarrow&                           SCE\textdownarrow&                           NLL\textdownarrow \\
\midrule
none& Beta &  \valmaxmin{13.55}{0.58}{0.48} & \valmaxmin{17.68}{0.81}{0.49} &  \valmaxmin{0.11}{0.02}{0.03} &  \valmaxmin{0.07}{0.03}{0.02} &  \valmaxmin{0.59}{0.04}{0.03} & \valmaxmin{0.07}{0.00}{0.00} \\
      
         \gls{nms} & none & \valmaxmin{40.76}{0.99}{0.79} &  \valmaxmin{61.10}{1.32}{1.26} &  \valmaxmin{7.82}{0.11}{0.07} &  
         \valmaxmin{7.82}{0.11}{0.07} &  \valmaxmin{7.69}{0.11}{0.10} &  \valmaxmin{0.21}{0.00}{0.01} \\
         \gls{nms} & Beta & \valmaxmin{40.76}{0.99}{0.79} &  \valmaxmin{61.10}{1.32}{1.26} &  \valmaxmin{0.22}{0.14}{0.13} &   
      \valmaxmin{0.21}{0.08}{0.08} &  \valmaxmin{1.97}{0.16}{0.10} &  \valmaxmin{0.17}{0.00}{0.01} \\
     
          soft & none & \valmaxmin{41.34}{1.01}{0.88} & \valmaxmin{61.18}{1.28}{1.20} & \valmaxmin{6.12}{0.06}{0.07} & \valmaxmin{6.12}{0.06}{0.07} & \valmaxmin{6.33}{0.09}{0.09} & \valmaxmin{0.17}{0.00}{0.00} \\
          soft & Beta &\valmaxmin{41.34}{1.01}{0.88} & \valmaxmin{61.18}{1.28}{1.20} & \valmaxmin{0.23}{0.05}{0.07} & \valmaxmin{0.22}{0.06}{0.05} & \valmaxmin{1.61}{0.11}{0.07} & \valmaxmin{0.13}{0.00}{0.00}  \\ 

      
         \gls{wbf} & none &\valmaxmin{37.50}{0.76}{0.70} &  \valmaxmin{58.00}{0.80}{0.88} &  \valmaxmin{4.47}{0.10}{0.12} &  
         \valmaxmin{4.41}{0.11}{0.12} &  \valmaxmin{4.64}{0.12}{0.15} &  \valmaxmin{0.19}{0.00}{0.01} \\
         \gls{wbf} & Beta & \valmaxmin{37.50}{0.76}{0.70} &  \valmaxmin{58.00}{0.80}{0.88} &  \valmaxmin{0.47}{0.09}{0.08} &   
      \valmaxmin{0.43}{0.08}{0.15} &  \valmaxmin{2.31}{0.20}{0.12} &  \valmaxmin{0.17}{0.00}{0.00} \\
none &  Cond.-Beta & \valmaxminb{41.36}{1.01}{0.85} &  \valmaxminb{61.28}{1.32}{1.23} &   \valmaxminb{0.03}{0.01}{0.01} &   
\valmaxminb{0.04}{0.02}{0.02} &   \valmaxminb{0.31}{0.02}{0.02} &  \valmaxminb{0.03}{0.00}{0.00} \\
\bottomrule
\end{tabular}}}%
        \vspace{1mm}%
        \caption{\textbf{Comparison of \method~to traditional \gls{nms} methods with calibration.} Averaged results with maximum positive and negative difference shown in brackets, lifted and lowered respectively. 
        }
    \label{tab:baseline}
\end{table}
\subsection{Ablations}\label{subsec:ablations}
Our \method~ delivers good results, which is a indication that our design- and modeling choices are well-founded and accurately reflect the underlying problem. Nonetheless, we evaluate the impact of each of the modeling choices and confirm their validity via ablations. 
\\
\shortsec{Choice of calibration function.} We chose the Beta calibration function over the Logistic function because of its higher expressiveness. We assumed a conditional dependence between the confidence \conf{i} of the detection and our proxy for the overlap with more confident detections, \jaccardd{i, \min}. Visual inspection of \cref{fig:compare_cc_softnms} confirms that the learned $\calibfunctrue$ models some interaction between the two variables. We compare the conditional dependent bi-variate Beta calibration to the independent version as well as the dependent and independent bi-variate Logistic calibration function. \\
As shown in \cref{tab:ablate_calib_func}, there are no significant differences between the calibration functions, except for the independent Logistic calibration, which performs significantly worse for both performance- and calibration-metrics. The independent Beta calibration has only a slightly higher \gls{ece} than its conditional counterpart. The performance differences between conditional Logistic and Beta calibrations are only minor. We use the Beta calibration for our experiments but note that a conditional Logistic calibration would likely produce similar results. \\
\begin{table}[t!] 
        \centering{%
        \resizebox{\columnwidth}{!}{%
        \begin{tabular}{lcccccc}
\toprule
Calibration-& \multicolumn{2}{c}{Performance-Metrics} & \multicolumn{4}{c}{Calibration-Metrics}  \\\cmidrule(lr){2-3}\cmidrule(lr){4-7}
Method &           \map \textuparrow &                          \mapfifty \textuparrow &                           ECE\textdownarrow &                           
                ACE\textdownarrow&                           SCE\textdownarrow&                           NLL\textdownarrow \\
\midrule
Ind.-Logistic & \valmaxmin{40.25}{0.94}{0.84} &  \valmaxmin{59.59}{1.22}{1.16} &   \valmaxmin{0.13}{0.03}{0.03} &   
\valmaxmin{0.14}{0.04}{0.03} &   \valmaxmin{0.46}{0.02}{0.02} &  \valmaxminb{0.03}{0.00}{0.00} \\

 Cond.-Logistic & \valmaxmin{41.33}{1.00}{0.81} &  \valmaxminb{61.42}{1.32}{1.26} &   \valmaxmin{0.04}{0.01}{0.01} &   
 \valmaxmin{0.05}{0.02}{0.02} &   \valmaxminb{0.31}{0.03}{0.02} &  \valmaxminb{0.03}{0.00}{0.00} \\
 
 Ind.-Beta & \valmaxmin{41.32}{0.98}{0.82} &  \valmaxmin{61.20}{1.31}{1.26} &   \valmaxmin{0.08}{0.01}{0.02} &   
 \valmaxminb{0.03}{0.02}{0.01} &   \valmaxmin{0.33}{0.02}{0.02} &  \valmaxminb{0.03}{0.00}{0.00} \\
 
   Cond.-Beta & \valmaxminb{41.36}{1.01}{0.85} &  \valmaxmin{61.28}{1.32}{1.23} &   \valmaxminb{0.03}{0.01}{0.01} &   
   \valmaxmin{0.04}{0.02}{0.02} &   \valmaxminb{0.31}{0.02}{0.02} &  \valmaxminb{0.03}{0.00}{0.00} \\
\bottomrule
\end{tabular}}}%
        \caption{\textbf{Ablation of different calibration methods.} 
        Conditionally dependent and independent versions of the Bi-variate Logistic- and Beta calibration are evaluated. 
        }
        \vspace{-0.5em}
    \label{tab:ablate_calib_func}
\end{table}
\shortsec{Conditioning on other summary-statistics.} We chose our summary statistic of \jaccardv{i} by taking the minimum Jaccard-distance \ie, the maximum \iou{}. The greedy \gls{nms} methods apply the confidence adjustments iteratively (see \cref{eq:discount_conf}), thereby accounting for individual overlapping detections, not only the extrema. This approach is not transferable to our calibration method, but we can look for a summary statistic that accounts for the additional overlapping detections. Ideally, the value should be bounded in [0, 1], making the product of the Jaccard-distances an ideal choice. We denote the resulting variable as \jaccardd{i, \Pi}.
We compute the statistics \jaccardd{i, \Pi} and \jaccardd{i, min} from \jaccardv{i}, over all the detections with \emph{higher} confidences than the detection \detect{i} that has its confidence \conf{i} adjusted. In the context of \gls{nms}, these would be the overlaps with \emph{suppressing} bounding boxes; this leaves the potential influence of boxes \emph{suppressed} by detection \detect{i} unaccounted for. To address this, we can also compute the proposed summary statistics for this, left out, suppressing part of the \jaccardv{i} and denote them as \jaccarddsup{i, min} and \jaccarddsup{i, \Pi} even though we have no prior reason to assume that these statistics are relevant for our calibration objective.
\\ 
Comparing the performance changes by conditioning on the different variables (see \cref{tab:ablate_singlevar}), we find that the conditioning on the minimum Jaccard-distance of the suppressing detections \jaccardd{\min} leads to significantly better performance and calibration scores than any of the alternatives. The introduced product \jaccardd{\Pi} is not as effective as \jaccardd{min} at capturing the likelihood of duplicate detections and the statistics over the suppressed detections, \jaccarddsup{min} and \jaccarddsup{\Pi}, are only slightly barely better at it than the baseline, which can be seen on the performance metrics. 
\begin{table}[t!] 
        \centering{%
        \resizebox{\columnwidth}{!}{%
        \begin{tabular}{lcccccc} 
\toprule
Conditioned- & \multicolumn{2}{c}{Performance-Metrics} & \multicolumn{4}{c}{Calibration-Metrics}  \\\cmidrule(lr){2-3}\cmidrule(lr){4-7}
                Variable &           \map \textuparrow &                          \mapfifty \textuparrow &                           ECE\textdownarrow &                           
                ACE\textdownarrow&                           SCE\textdownarrow&                           NLL\textdownarrow \\
\midrule
\jaccardd{\min} & \valmaxminb{41.34}{1.02}{0.81} & \valmaxminb{61.26}{1.34}{1.20} &  \valmaxminb{0.04}{0.01}{0.01} &   
\valmaxminb{0.04}{0.02}{0.02} &  \valmaxminb{0.31}{0.03}{0.02} & \valmaxminb{0.03}{0.00}{0.00}  \\
\jaccardd{\Pi} & \valmaxmin{40.50}{0.81}{0.89} & \valmaxmin{59.35}{1.10}{1.17} &  \valmaxmin{0.06}{0.03}{0.03} &   
\valmaxmin{0.05}{0.02}{0.03} &  \valmaxmin{0.37}{0.04}{0.02} & \valmaxminb{0.03}{0.00}{0.00}  \\
\jaccarddsup{\min} & \valmaxmin{14.48}{0.59}{0.60} & \valmaxmin{19.26}{0.77}{0.72} &  \valmaxmin{0.10}{0.02}{0.03} &  
\valmaxmin{0.08}{0.02}{0.02} &  \valmaxmin{0.60}{0.04}{0.03} & \valmaxmin{0.07}{0.00}{0.00} \\
\jaccarddsup{\Pi} & \valmaxmin{13.72}{0.77}{0.54} & \valmaxmin{17.95}{0.98}{0.62} &  \valmaxmin{0.08}{0.18}{0.06} & 
\valmaxmin{0.09}{0.20}{0.05} &  \valmaxmin{0.56}{0.09}{0.03} & \valmaxmin{0.07}{0.00}{0.00}  \\
none &\valmaxmin{13.55}{0.58}{0.48} & \valmaxmin{17.68}{0.81}{0.49} &  \valmaxmin{0.11}{0.02}{0.03} & 
\valmaxmin{0.07}{0.03}{0.02} &  \valmaxmin{0.59}{0.04}{0.03} & \valmaxmin{0.07}{0.00}{0.00} \\
\bottomrule
\end{tabular}}}%
        \caption{\textbf{Ablation conditional calibration variables.} Comparing the performance by taking summary statistics over all the \emph{suppressing} detections (\jaccardd{\Pi} and \jaccardd{\min}) and over the detections that are being \emph{suppressed} (\jaccarddsup{\min} and \jaccarddsup{\Pi}). 
        }%
    \label{tab:ablate_singlevar}
\end{table}
%
\begin{table}[t] 
        \centering{%
        \resizebox{\columnwidth}{!}{%
        \begin{tabular}{ccccccccc}
\toprule
\multicolumn{4}{c}{Variables} & \multicolumn{2}{c}{Performance-Metrics} & \multicolumn{3}{c}{Calibration-Metrics}  \\\cmidrule(lr){1-4}\cmidrule(lr){5-6}\cmidrule(lr){7-9}
    \small{\jaccardd{\min}} & \small{\jaccardd{\Pi}} & \small{\jaccarddsup{\min}} & \small{\jaccarddsup{\Pi}} &                       \map \textuparrow &                          \mapfifty \textuparrow &                           ECE\textdownarrow &                           
                ACE\textdownarrow&                           SCE\textdownarrow                           
                \\
\midrule
                                \cmark &                         &                     &                          &\valmaxmin{41.21}{1.18}{1.39} & \valmaxmin{61.07}{1.55}{1.85} & \valmaxmin{0.05}{0.16}{0.03} &   
                                \valmaxmin{0.04}{0.12}{0.04} &  \valmaxmin{0.32}{0.09}{0.03}  
                                \\
            \cmark &        \cmark &                         &                          & \valmaxmin{41.33}{1.07}{0.81} & \valmaxmin{61.21}{1.41}{1.57} &  \valmaxmin{0.04}{0.03}{0.01} &  
\valmaxmin{0.03}{0.04}{0.02} &  \valmaxmin{0.31}{0.03}{0.02}  
\\
           \cmark &                      &            \cmark &                          & \valmaxmin{41.14}{1.23}{1.17} & \valmaxmin{60.97}{1.61}{1.39} &  \valmaxmin{0.06}{0.15}{0.04} &  
\valmaxmin{0.05}{0.11}{0.04} &  \valmaxmin{0.32}{0.08}{0.03}  
\\
       \cmark &        \cmark &            \cmark &                          & \valmaxmin{41.22}{1.15}{0.99} & \valmaxmin{61.06}{1.52}{1.29} &  \valmaxmin{0.05}{0.12}{0.02} &  
\valmaxmin{0.04}{0.05}{0.02} &  \valmaxmin{0.32}{0.05}{0.02}  
\\
        \cmark &                     &                         &             \cmark & \valmaxmin{41.14}{1.01}{1.31} & \valmaxmin{60.97}{1.39}{1.75} &  \valmaxmin{0.07}{0.10}{0.03} &  
\valmaxmin{0.05}{0.12}{0.03} &  \valmaxmin{0.32}{0.04}{0.03}  
\\
      \cmark &                     &            \cmark &             \cmark & \valmaxmin{41.09}{1.21}{1.12} & \valmaxmin{60.93}{1.60}{1.31} &  \valmaxmin{0.08}{0.13}{0.06} &  
\valmaxmin{0.05}{0.11}{0.02} &  \valmaxmin{0.33}{0.08}{0.03}  
\\
     \cmark &        \cmark &                         &             \cmark & \valmaxmin{41.28}{1.08}{0.75} & \valmaxmin{61.13}{1.43}{1.49} &  \valmaxmin{0.04}{0.03}{0.01} &  
\valmaxmin{0.03}{0.04}{0.03} &  \valmaxmin{0.31}{0.02}{0.02}  
\\
       \cmark &        \cmark &            \cmark &             \cmark & \valmaxmin{41.27}{1.07}{0.76} & \valmaxmin{61.11}{1.45}{1.15} &  \valmaxmin{0.04}{0.06}{0.02} &  
\valmaxmin{0.03}{0.01}{0.01} &  \valmaxmin{0.31}{0.05}{0.02}  
\\

\bottomrule
\end{tabular}
}}%
        \caption{\textbf{Ablation of combination of additional calibration variables.} We apply conditional multivariate calibration with \jaccardd{\min} and different combination of the additional summary statistics (\jaccardd{\Pi}, \jaccarddsup{\min} and \jaccarddsup{\Pi}).
        }
    \label{tab:ablate_variate_comb}
\end{table}
\begin{table}[t]
        \centering{%
        \resizebox{\columnwidth}{!}{%
        \begin{tabular}{llcc}
\toprule
Object Detector & Post-Processing &                          \map\textuparrow &                          \mapfifty\textuparrow  \\
\midrule
Sparse-RCNN Rn50 \cite{sun2021sparsercnn} &
none (default) &  \valmaxmin{45.49}{1.10}{1.33} &  \valmaxmin{64.38}{1.28}{1.51} \\
& \gls{nms} &  \valmaxminb{45.67}{1.11}{1.33} &  \valmaxminb{64.77}{1.30}{1.50} 
\\
&   cond. Beta & \valmaxmin{45.48}{1.10}{1.33} &  \valmaxmin{64.37}{1.28}{1.51} 
\\
\hline
CenterNet HG \cite{zhou2019objects} 
& none (default) & \valmaxmin{40.66}{0.53}{0.84} & \valmaxmin{59.28}{1.11}{1.28} \\
& \gls{nms} & \valmaxmin{40.73}{0.52}{0.83} & \valmaxmin{59.45}{1.13}{1.28} 
\\
& cond. Beta & \valmaxminb{40.76}{0.57}{0.82} & \valmaxminb{59.61}{1.14}{1.29} 
\\
\hline
Detr Rn50 \cite{carion2020endtoend_detr} 
& none (default) & \valmaxmin{40.49}{0.74}{1.10} & \valmaxmin{60.89}{1.29}{1.68} \\
& \gls{nms} & \valmaxmin{40.56}{0.74}{1.08} & \valmaxmin{61.11}{1.29}{1.65} 
\\

& cond. Beta & \valmaxminb{40.64}{0.73}{1.07} &  \valmaxminb{61.62}{1.23}{1.70} 
\\
\bottomrule
\end{tabular}}}%
        \caption{\textbf{We compare post-processing methods on Detection architectures designed for No-\gls{nms}.} As a sanity check we evaluate the \method on architectures that do not require \gls{nms}. 
        }
        \vspace{-0.5em}
    \label{tab:no_nms}
\end{table}
\\
\shortsec{Conditioning on additional variables.} Up to this point, we only performed Bi-variate calibration using the confidence \conf{} and a single summary statistic over all Jaccard-distances to the other detections $(\jaccardv{i})$. While the other proposed metrics did not perform well on their own, we could combine them with the best statistic, \jaccardd{min}, to a multivariate confidence calibration to see if jointly they are better able to model the likelihood of duplication. \\
We explore all possible combinations of the other available summary statistics, \jaccardd{\Pi}, \jaccarddsup{\min} and \jaccarddsup{\Pi} with \jaccardd{\min} and \conf{} (see \cref{tab:ablate_variate_comb}). 
There is no significant improvement, as seen in the performance and calibration metrics. Combinations that include \jaccardd{\Pi} seem to have a slight edge over the ones that do not include it. However, unlike the contribution of conditioning on \jaccardd{\min}, these gains are far from significant enough to justify modeling the conditioning on an additional variable.  
\begin{table*}[ht!]
        \centering{%
        \resizebox{2\columnwidth}{!}{%
        \begin{tabular}{lllcccccccc}
\toprule
Object-&\multicolumn{2}{c}{Post-Processing}& \multicolumn{2}{c}{\texttt{test-dev2017}} &\multicolumn{6}{c}{\texttt{val2017} (Cross-Val)}   \\\cmidrule(lr){2-3}\cmidrule(lr){4-5}\cmidrule(lr){6-11}
Detector &   \gls{nms} & Calibration &                          \map\textuparrow &                          \mapfifty\textuparrow &                           ECE\textdownarrow &                          
ACE\textdownarrow&                           SCE\textdownarrow&                           NLL\textdownarrow &            \map\textuparrow &                          \mapfifty\textuparrow \\
\midrule
YoloV3-608 \cite{redmon2018yolov3} &\cmark & none &    34.2  & 56.4 &
\valmaxmin{34.78}{0.87}{0.61} & \valmaxmin{57.16}{1.16}{1.17} & \valmaxmin{1.55}{0.09}{0.12} & \valmaxmin{1.31}{0.13}{0.10} & \valmaxmin{2.95}{0.14}{0.15} & \valmaxmin{0.19}{0.00}{0.00} 
\\
&\cmark &  Beta &  34.2  & 56.4 &
\valmaxmin{34.78}{0.87}{0.61} & \valmaxmin{57.16}{1.16}{1.17} & \valmaxmin{0.28}{0.05}{0.07} & \valmaxmin{0.25}{0.10}{0.10} & \valmaxmin{2.71}{0.13}{0.18} & \valmaxmin{0.19}{0.00}{0.00} 
 \\
& \xmark &  cond. Beta &  \best{34.6} & \best{56.7} &
\valmaxminb{35.28}{0.87}{0.54} & \valmaxminb{57.62}{1.16}{1.17} & \valmaxminb{0.22}{0.12}{0.05} & \valmaxminb{0.13}{0.07}{0.07} & \valmaxminb{1.22}{0.09}{0.09} & \valmaxminb{0.09}{0.00}{0.00} 
\\
\hline
RetinaNet Rn101 \cite{lin2017focal} &\cmark & none &  \best{38.8} & \best{56.3} &
\valmaxminb{38.71}{0.67}{1.18} & \valmaxminb{55.50}{0.93}{1.28} & \valmaxmin{6.85}{0.06}{0.05} & \valmaxmin{6.86}{0.09}{0.07} & \valmaxmin{7.49}{0.10}{0.10} & \valmaxmin{0.17}{0.00}{0.00} 
\\
&\cmark &  Beta &  \best{38.8} & \best{56.3} &
\valmaxminb{38.71}{0.67}{1.18} & \valmaxminb{55.50}{0.93}{1.28} & \valmaxmin{0.19}{0.12}{0.10} & \valmaxmin{0.20}{0.09}{0.12} & \valmaxmin{2.00}{0.14}{0.14} & \valmaxmin{0.12}{0.00}{0.00}
\\
 &  \xmark &  cond. Beta &  {38.6} & {56.0} &
 \valmaxmin{38.64}{0.68}{1.14} & \valmaxmin{55.31}{0.87}{1.28} &  \valmaxminb{0.06}{0.02}{0.02} &  \valmaxminb{0.04}{0.03}{0.02} &  \valmaxminb{0.28}{0.02}{0.01} & \valmaxminb{0.02}{0.00}{0.00} 
\\
\hline
Faster-RCNN Rn50 \cite{Ren2017fasterrcnn} &\cmark & none &  \best{41.4}  & 61.6 &
\valmaxmin{41.34}{1.01}{0.88} & \valmaxmin{61.18}{1.28}{1.20} & \valmaxmin{6.12}{0.06}{0.07} & \valmaxmin{6.12}{0.06}{0.07} & \valmaxmin{6.33}{0.09}{0.09} & \valmaxmin{0.17}{0.00}{0.00}
 \\

 &\cmark &  Beta &  \best{41.4} & 61.6 &
 \valmaxmin{41.34}{1.01}{0.88} & \valmaxmin{61.18}{1.28}{1.20} & \valmaxmin{0.23}{0.05}{0.07} & \valmaxmin{0.22}{0.06}{0.05} & \valmaxmin{1.61}{0.11}{0.07} & \valmaxmin{0.13}{0.00}{0.00} 
 \\
 &  \xmark &  cond. Beta & {41.3} & \best{61.8} &
 \valmaxminb{41.35}{1.01}{0.82} & \valmaxminb{61.28}{1.32}{1.22} &    \valmaxminb{0.04}{0.01}{0.01} &  \valmaxminb{0.04}{0.02}{0.02} &  \valmaxminb{0.31}{0.03}{0.02} & \valmaxminb{0.03}{0.00}{0.00} 
 \\
 \hline
Varifocalnet Rn50 \cite{zhang2021varifocalnet}  & \cmark & none &  47.7 & \best{65.8} & \valmaxmin{47.85}{0.61}{1.00} & \valmaxminb{65.65}{0.97}{1.39} & \valmaxmin{6.47}{0.03}{0.06} & \valmaxmin{5.77}{0.06}{0.08} & \valmaxmin{6.92}{0.10}{0.03} & \valmaxmin{0.14}{0.00}{0.00} 
\\
 & \cmark & Beta &  47.7 & \best{65.8} &
 \valmaxmin{47.85}{0.61}{1.00} & \valmaxminb{65.65}{0.97}{1.39} & \valmaxmin{0.15}{0.03}{0.04} & \valmaxmin{0.10}{0.04}{0.03} & \valmaxmin{1.31}{0.07}{0.07} & \valmaxmin{0.10}{0.00}{0.00}
 \\
 & \xmark &  cond. Beta &  \best{47.9} & {65.7} &
 \valmaxminb{48.16}{0.63}{1.02} &  \valmaxmin{65.59}{0.97}{1.39} &   \valmaxminb{0.09}{0.02}{0.03} &  \valmaxminb{0.05}{0.01}{0.01} &   \valmaxminb{0.35}{0.03}{0.02} &  \valmaxminb{0.03}{0.00}{0.00}  
\\
\hline
YOLOX-L \cite{ge2021yolox}  &\cmark & none &  49.3 & 67.5 &
\valmaxmin{49.44}{0.89}{1.72} &  \valmaxmin{67.22}{1.41}{1.95} &  \valmaxmin{1.83}{0.11}{0.14} &   \valmaxmin{1.80}{0.10}{0.10} &  \valmaxmin{4.55}{0.26}{0.23} &  \valmaxmin{0.23}{0.00}{0.00} 
 \\
 &\cmark & Beta &   49.3 & 67.5&
 \valmaxmin{49.44}{0.89}{1.72} &  \valmaxmin{67.22}{1.41}{1.95} &  \valmaxmin{0.26}{0.12}{0.06} &  \valmaxmin{0.27}{0.11}{0.13} &  \valmaxmin{3.99}{0.23}{0.17} &  \valmaxmin{0.22}{0.00}{0.00} 
 \\
 & \xmark &  cond. Beta &  \best{49.5} & \best{67.8} &
 \valmaxminb{49.69}{0.96}{1.65} &  \valmaxminb{67.50}{1.33}{1.91} &  \valmaxminb{0.19}{0.03}{0.03} &   \valmaxminb{0.08}{0.07}{0.03} &   \valmaxminb{0.94}{0.05}{0.07} &  \valmaxminb{0.06}{0.00}{0.00}  
\\
\hline
HTC CBNetv2 Swin-L $\dagger$ \cite{liang2022cbnet}  &\cmark & none &  58.4 & 76.0 & \valmaxmin{58.56}{0.60}{1.55} & \valmaxmin{75.75}{0.81}{1.70} & \valmaxmin{8.63}{0.33}{0.30} & \valmaxmin{8.63}{0.33}{0.30} & \valmaxmin{9.13}{0.25}{0.51} & \valmaxmin{0.26}{0.01}{0.01} 
\\
&\cmark & Beta &   58.4 & 76.0 &
\valmaxmin{58.56}{0.60}{1.55} & \valmaxmin{75.75}{0.81}{1.70} & \valmaxmin{0.38}{0.10}{0.12} & \valmaxmin{0.35}{0.19}{0.14} & \valmaxmin{4.10}{0.15}{0.34} & \valmaxmin{0.21}{0.01}{0.01} 
\\
& \xmark & cond. Beta &    \best{58.8} &  \best{76.8} &
\valmaxminb{58.99}{0.64}{1.44} & \valmaxminb{76.45}{0.89}{1.59} &  \valmaxminb{0.06}{0.03}{0.02} &  \valmaxminb{0.07}{0.05}{0.03} &  \valmaxminb{0.62}{0.02}{0.05} & \valmaxminb{0.03}{0.00}{0.00} 
\\

\hline
 EVA Cascade Mask-RCNN $\dagger$ \cite{fang2022eva} &\cmark & none &  \best{63.0} & 81.6 & \valmaxminb{63.07}{0.83}{1.11} & \valmaxminb{81.56}{0.52}{1.09} & \valmaxmin{1.21}{0.07}{0.05} & \valmaxmin{1.21}{0.07}{0.05} & \valmaxmin{1.47}{0.06}{0.08} & \valmaxmin{0.08}{0.00}{0.00} 
\\
 & \cmark &  Beta &  \best{63.0} & 81.6 &
 \valmaxminb{63.07}{0.83}{1.11} & \valmaxminb{81.56}{0.52}{1.09} & \valmaxminb{0.09}{0.03}{0.02} & \valmaxminb{0.06}{0.05}{0.04} & \valmaxminb{0.98}{0.04}{0.05} & \valmaxminb{0.07}{0.00}{0.00} 
 \\
 &  \xmark &  cond. Beta &  \best{63.0} & \best{81.9} &
 \valmaxminb{63.07}{0.85}{1.11} & \valmaxminb{81.56}{0.51}{1.10} & \valmaxminb{0.09}{0.03}{0.01} &   \valmaxmin{0.07}{0.02}{0.02} & \valmaxminb{0.98}{0.04}{0.06} & \valmaxminb{0.07}{0.00}{0.00} 
 \\
\bottomrule
\end{tabular}
        }}%
        \caption{\textbf{Proposed \method~is applied on a variety of Object detectors.} We apply our conditional calibration to a wide range of object detectors, including state-of-the-art detectors, and compare it to the best (oracle) \gls{nms} alternative. We evaluate the detections for performance- and calibration-metrics on 10 random splits on the validation set and report the mean and maximum deviation. On the test-dev set, we show only performance-metrics produced by the evaluation server, as ground truth labels are not publicly available. Detectors marked with $\dagger$ are also instance-segmentation models. 
        } 
    \label{tab:apply_all_dets}
\end{table*}
\subsection{Results}
Finally, we verify the effectiveness of our \method~on a wide range of object detection architectures. We aim to get representatives of the different design philosophies: from single-stage to two-stage, from small, real-time to huge billion-parameter models, anchor-based and anchor-free, designed for \gls{nms} or without, transformer- and CNN-based models, and different backbone-networks. The results on all object detectors are shown in \cref{tab:apply_all_dets}. As a sanity check we also verify that there is little performance variation on architectures that are designed without \gls{nms} post-processing (see \cref{tab:no_nms}). The method consistently outperforms standard \gls{nms}---for some models by up to 0.7 \map{}---and can often improve on the oracle-tuned soft-\gls{nms} by up to 0.4 \map{}.
\section{Discussion}
The performance of \method~is on-par with with the best fine-tuned \gls{nms}-method across models which proves that it is able to model the underlying problem of duplicate detections regardless of architecture and better than any individual type of \gls{nms}. The better-calibrated confidence predictions further show that the calibrations is not only good at implicitly capturing the likelihood of duplication, but the duplication likelihood is also a crucial intermediary for confidence predictions that accurately reflect the empirical precision. This shows that we can fuse the de-duplication into the calibration into a single step, simplifying the post-processing and improving calibration and performance with reduced complexity.
%
\\
\shortsec{Limitations.} The \method~is a data-driven approach, so a distribution shift between the data used for calibration and data seen at deployment can lead to over- or under-confident predictions. It can also lead to performance degradation in significantly more crowded scenes than the calibration subset. 
\section{Conclusion}
We proposed \method~as a data-driven alternative to the classic iterative \gls{nms} and calibration post-processing. We showed how our method implicitly accounts for the likelihood of each detection being a duplicate of another and adjusts the confidence score accordingly. This results in well-calibrated probability estimates for the precision of each detection. With comprehensive ablation studies, we demonstrated the validity of our design choices. Furthermore, we conducted extensive experiments across various detection architectures, we found that the proposed IoU-aware Calibration method results in performance gains comparable to those obtained by an oracle chosen fine-tuned non-maximum suppression algorithm. In some cases, it even achieves performance gains over the best \gls{nms}-based alternative while also producing consistently better calibrated confidence predictions than comparable calibrations. We see this as a significant step away from opaque hand-crafted algorithms toward data-driven detectors with interpretable outputs.

{\small
\bibliographystyle{ieee_fullname}
\bibliography{paper_bib}
}

\clearpage
\appendix
\section{Influence of \iouthresh{} on \method{}}
For fitting the proposed \method{} we need to evaluate the detections of a object detector. The concept of a \gls{tp} for detectors is more involved than for \eg classification, as it depends on the chosen \iouthresh{} which defines the minimum overlap required for a detection with an actual object to be considered a \gls{tp} detection. As mentioned in the Background section, in our experiments we followed Küppers \etal \cite{kuppers2020multivariate} and use a \iou{} threshold \iouthresh{} of 0.5. The same threshold is used for fitting the \method{} and for evaluating how well the detections are calibrated. The choice of \iouthresh{} can impact the performance changes of conditional confidence calibrations \cite{gilg2023box}. 
\vspace{0.5em}\\
\subsection{Impact on performance.} ]
In \cref{tab:iou_thes} we can see the impact of \iouthresh{} on the performance metrics. A \iouthresh{} of 0.5 makes the concept of a \gls{tp} the same for the calibration objective as it is for the evaluation metric \mapfifty{}, so unsurprisingly \mapfifty is maximized for \iouthresh{}=0.5. A \iouthresh{} of 0.60 leads to a slightly higher \map, but also a reduced \mapfifty. There is a severe performance drop for \iouthresh{}=0.9. This drop goes hand in hand with a sharp drop in the number of \gls{tp} targets \tpindic{} which is also likely a part of the reason for the performance drop. The smaller number of \gls{tp} detections makes it harder to properly fit the the calibration curve and can introduce artifacts from outliers in low density regions.  
\vspace{0.5em}\\
\subsection{Impact on calibration curve.} 
\begin{figure}[!]
        \centering{%
        \resizebox{0.9\columnwidth}{!}{%
        \includegraphics{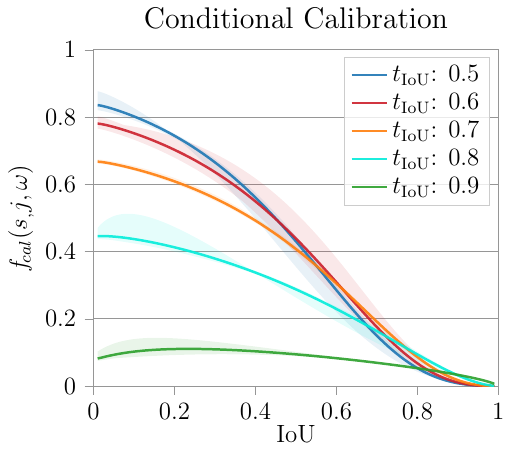}}}%
        \vspace{1mm}%
        \caption{\textbf{Comparison of the impact of \iouthresh{} for TP detections on the conditional calibration curves.} Shows how confidence of detections is adjusted, depending on the \iou{} with a more confident detection with initial confidences \conf{}=0.9. Confidence intervals in lighter colours.
        }
    \label{fig:iou_thes_curves}
\end{figure}

In \cref{fig:iou_thes_curves} we plotted the calibration curves for the range of \iouthresh{} values and an initial confidence of 0.9. Here we can also observe that \iouthresh{}=0.9 breaks the trend of the other thresholds and bends lower for very small \iou{} values. This is likely an artifact of the Beta calibration function. 
\begin{table}[!]
        \centering{%
        \begin{tabular}{lccc}
\toprule
\iouthresh &    \# \tpindic  &                      \map\textuparrow &                          \mapfifty\textuparrow  \\
\midrule
0.50 & 31282 & \valmaxmin{41.36}{1.00}{0.82} & \valmaxminb{61.28}{1.32}{1.23} \\
0.60 & 28068 & \valmaxminb{41.40}{1.01}{0.82} & \valmaxmin{61.02}{1.33}{1.17}\\
0.70 & 26795 & \valmaxmin{41.32}{1.02}{0.88} & \valmaxmin{60.47}{1.34}{1.18} \\
0.80 & 23855 & \valmaxmin{40.84}{1.03}{0.87} & \valmaxmin{59.11}{1.31}{1.22} \\
0.90 & 15933 & \valmaxmin{37.70}{0.94}{1.00} & \valmaxmin{53.23}{1.13}{1.53} \\
\bottomrule
\end{tabular}}
        \vspace{1mm}%
        \caption{\textbf{Comparison of the impact \iouthresh{} on the performance of \method{}.} We vary the \iouthresh{} that used to determine \tpindic{}---the optimization target for our conditional confidence calibration---from 0.5 to 0.9.  
        }
    \label{tab:iou_thes}
\end{table}
\begin{table}[!]
        \centering{%
        \resizebox{0.99\columnwidth}{!}{%
        \begin{tabular}{llcccc}
\toprule
NMS-Type & parameter & interval start & interval stop & spacing &   steps  \\
\midrule
\gls{nms} & \nmsthres & 0.40 & 0.90 & linear & 11 \\
Soft-\gls{nms}\cite{bodla2017soft_nms}& $\sigma$ & 0.001 & 0.20 & log & 20 \\
\gls{wbf} & \nmsthres & 0.50 & 0.90 & linear & 11 \\
\bottomrule
\end{tabular}
}}%
        \vspace{1mm}%
        \caption{\textbf{Settings for \gls{nms} hyper-parameter sweep.}
        }
        \vspace{-2mm}%
    \label{tab:nms_sweep}
\end{table}
\begin{figure*}[t!]%
    \centering{
        \begin{subfigure}[]{0.45\linewidth}%
        \includegraphics[width=\linewidth]{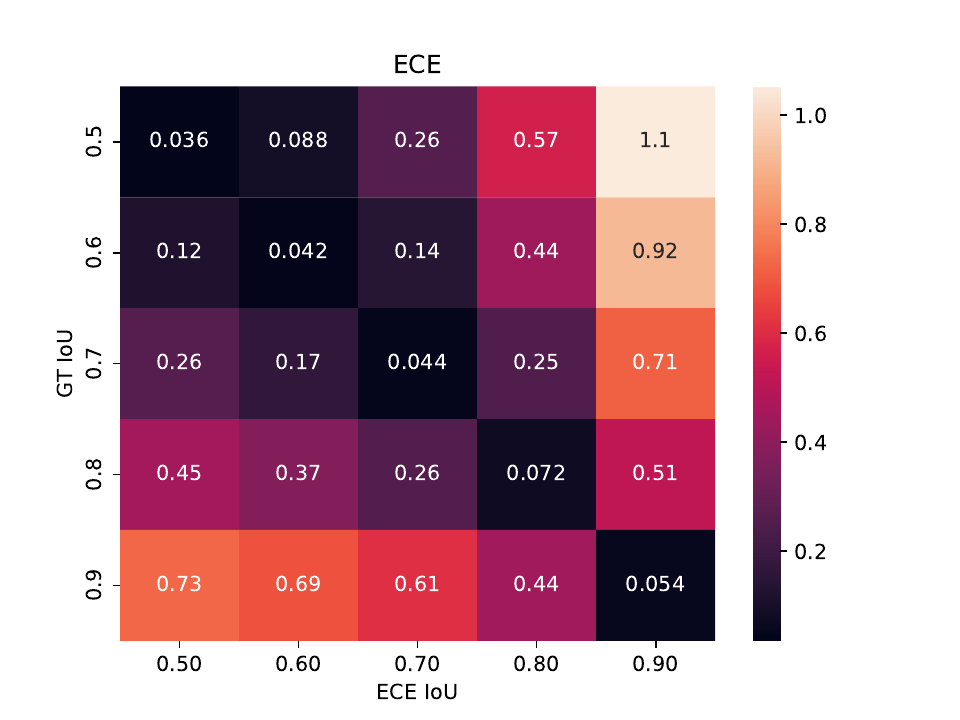}
        \caption{\quad\quad\quad\quad}
        \label{fig:iou_ece}%
        \end{subfigure}%
        \begin{subfigure}[]{0.45\linewidth}{%
        \includegraphics[width=\linewidth]{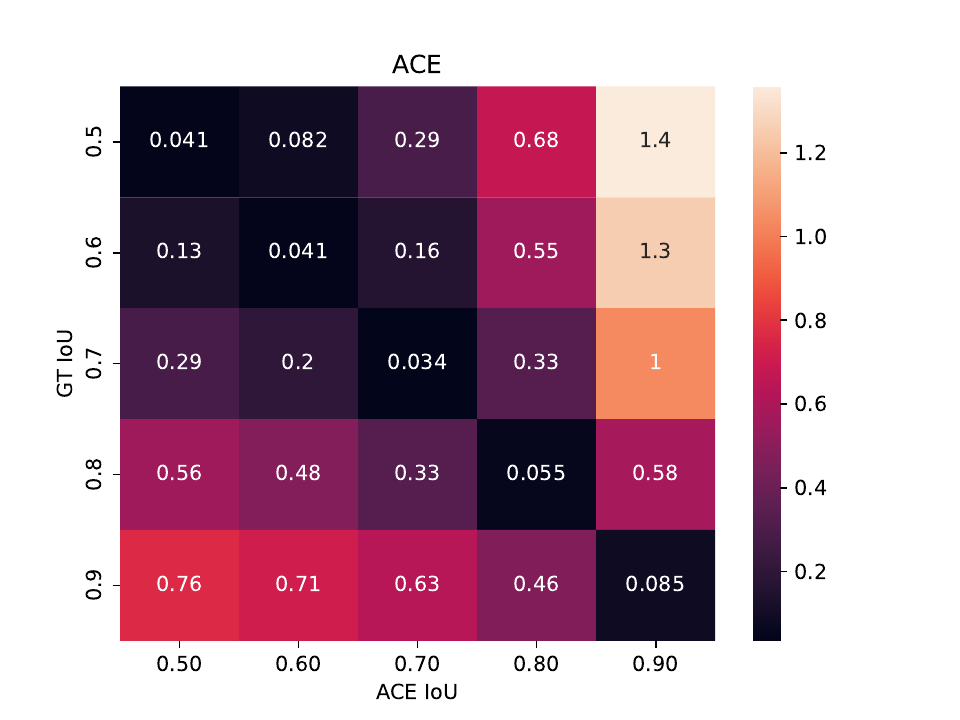}
        \caption{\quad\quad\quad\quad}
        \label{fig:iou_ace}}
        \end{subfigure}%
        
        \begin{subfigure}[]{0.45\linewidth}{%
        \includegraphics[width=\linewidth]{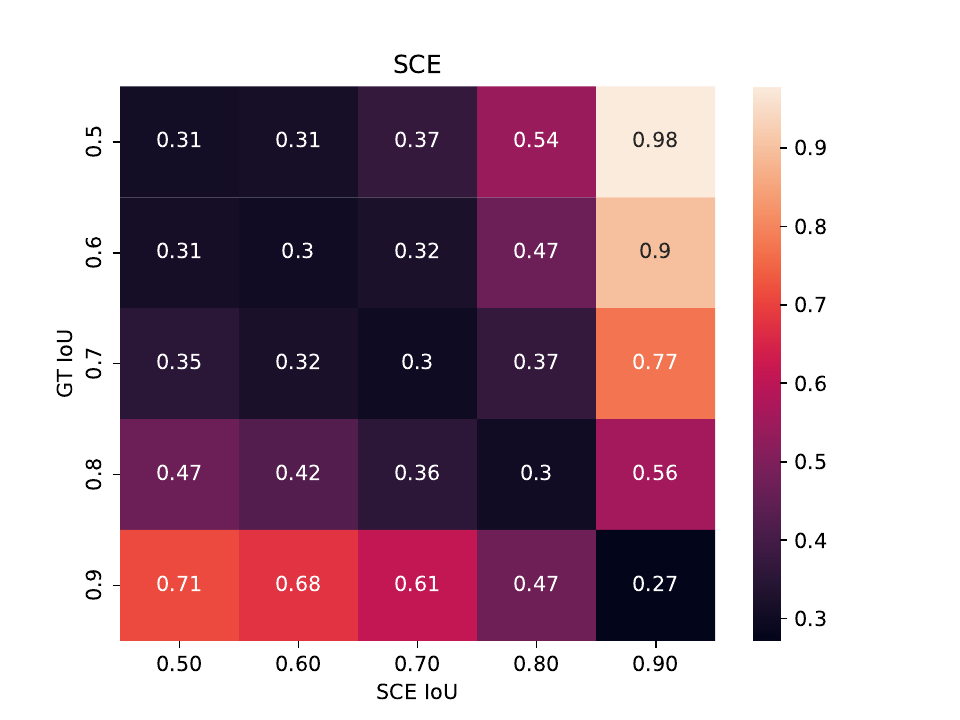}
        \caption{\quad\quad\quad\quad}
        \label{fig:iou_sce}}%
        \end{subfigure}%
        \begin{subfigure}[]{0.45\linewidth}{%
        \includegraphics[width=\linewidth]{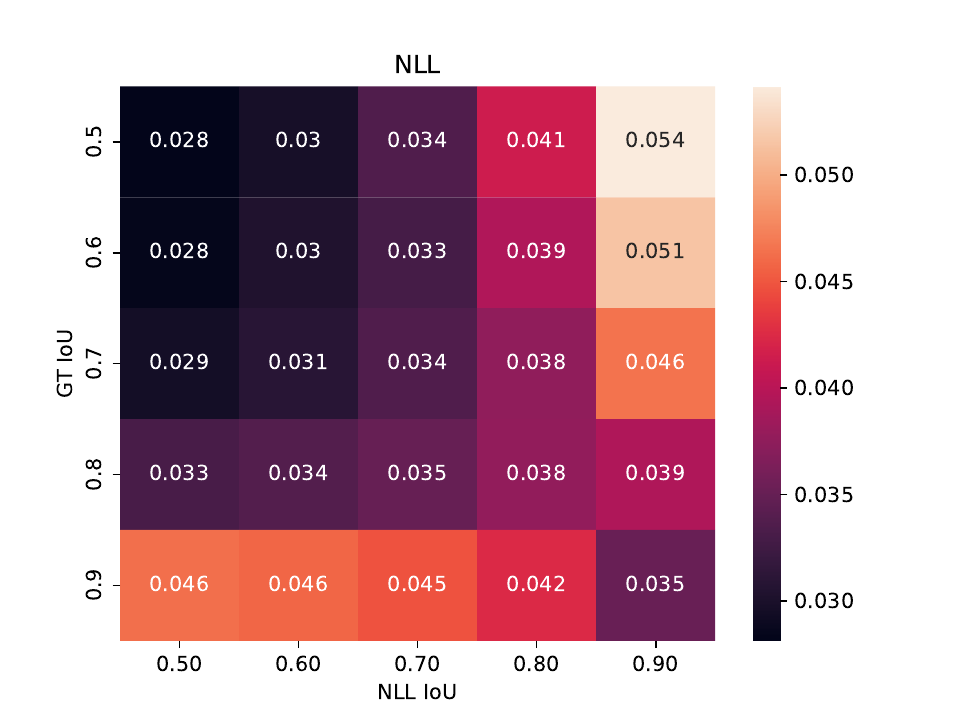}
        \caption{\quad\quad\quad\quad}
        \label{fig:iou_nll}}%
        \end{subfigure}%
        \caption{\textbf{Comparison of \iouthresh~values required for a detection to be considered a \gls{tp} detection.} On the Y-axis the \iouthresh~for the labels used for fitting the conditional confidence calibration is varied from 0.5 to 0.9, on the X-axis the corresponding \iouthresh~for the labels used for the calibration metric is varied from 0.5 to 0.9. The evaluated calibration metrics are (a) \gls{ece}, (b) \gls{ace}, (c) \gls{sce}, and (d) \gls{nll}.}
        \label{fig:iou_thres}
    }
\end{figure*}
\begin{table*}[t!]
        \centering{%
        \resizebox{0.99\linewidth}{!}{%
        \begin{tabular}{lllcccc}
\toprule
Model & Backbone & Settings &    Default NMS & Best NMS & Reported \map & Used implem. \map  \\
\midrule
Varifocalnet Rn50 \cite{zhang2021varifocalnet}  & ResNet-50 & e:24, DCNv2, FPN & \nmsthres= 0.60 & $\sigma$= 0.6 & 44.3 & 47.8 \\
YOLOX-L \cite{ge2021yolox} & CSP-V5 & e:300 & \nmsthres= 0.65 & \nmsthres=0.7 & 50.0 & 49.4 \\
Faster-RCNN Rn50 \cite{Ren2017fasterrcnn} & ResNet-50 & e:36, FPN, MS & \nmsthres= 0.70 & $\sigma$= 0.5 & - & 40.3 \\
YoloV3-608 \cite{redmon2018yolov3} &  DarkNet-53 & e:273 & \nmsthres= 0.45 & $\sigma$= 0.3 & 33.0 & 33.7 \\
RetinaNet Rn101 \cite{lin2017focal} &  ResNet-50 & e:24, MS, FPN & \nmsthres= 0.50 & $\sigma$=0.6 & 37.8 & 38.9 \\
\midrule
HTC CBNetv2 Swin-L $\dagger$ \cite{liang2022cbnet} & Swin-L & e:12, MS & $\sigma$ = 0.001 & $\sigma$=0.4 & 59.1 & 59.1  \\
EVA Cascade Mask-RCNN  $\dagger$ \cite{fang2022eva} & EVA & e:~24 & \nmsthres= 0.60 & \nmsthres= 0.50 & 64.1 & 63.9  \\
\midrule
Sparse-RCNN Rn50 \cite{sun2021sparsercnn} &  ResNet-50 & e:36, FPN, MS & none & \nmsthres= 0.80 & 45.0 & 45.0 \\
CenterNet HG \cite{zhou2019objects} &  Hourglass-104 & e:50 & none & \nmsthres= 0.80 & 42.1 & 40.3  \\
Detr Rn50 \cite{carion2020endtoend_detr} &  ResNet-50 & e:150, DCNv2 & none & \nmsthres= 0.85 & 42.0 & 40.1 \\
\bottomrule
\end{tabular}
}}%
        \vspace{1mm}%
        \caption{\textbf{Settings for all detectors.} Abbreviations: e refers to the number of training epochs, FPN means the Feature Pyramid Network Neck \cite{lin2017featurefpn} is used, MS is multi-scale training, and DCN indicates that Deformable Convolutions \cite{dai2017deformable} are used. $\sigma$ is the hyper-parameter for Gaussian Soft-NMS and \nmsthres{} the hard threshold for NMS. ``Reported \map{}" refers to the \map{} value reported in the publication that introduced the relevant model. ``Used implem." \map{} on the other hand refers to the performance of the model implementation we used for our experiments. 
        \vspace{-2mm}}
    \label{tab:settings_dets}
\end{table*}
\subsection{Varying \iouthresh{} for the calibration metrics.} 
Same as with variation of the \iouthresh{} for TP detections of the conditional calibration we can also change the \iouthresh{} for the calibration metrics. We show a grid of the resulting calibration metrics in \cref{fig:iou_thres}. The \gls{ece}, \gls{ace}, and \gls{sce} all follow a similar trend: the respective calibration metric is minimized for if the fitting- and the metric-\iouthresh{} are the same. There is, again, a sharp drop-off if one of the \iouthresh{} values is 0.9 and the other is not. Otherwise the miscalibration increases with increased distance between the two \iouthresh{} values.  


\section{Detector Architectures}
See \cref{tab:settings_dets} for more detailed settings on the used detection architectures and the best found hyper parameters for NMS. In \cref{tab:nms_sweep} we list the intervals for the \gls{nms} hyper-parameter sweep for the detectors. 

\end{document}